\documentclass[letterpaper, 10 pt, conference]{ieeeconf}  
\pdfoutput=1

\IEEEoverridecommandlockouts                              

\usepackage{graphicx} 
\usepackage{amsmath} 
\usepackage{amssymb}  
\usepackage{float}
\usepackage{xcolor}
\usepackage[caption=false, font=footnotesize]{subfig} 
\let\labelindent\relax
\usepackage{enumitem}
\usepackage{siunitx}

\usepackage[font=footnotesize]{caption}
\usepackage{titlesec}
\usepackage{adjustbox}

\usepackage{tabularx,booktabs}

\usepackage{titlesec}
\titlespacing{\section}{0pt}{*0.1}{*0.1}
\titlespacing{\subsection}{2pt}{*0.2}{*0.2}
\usepackage{todonotes}

\title{\LARGE \bf
Passing Through Narrow Gaps with Deep Reinforcement Learning
}

\author{Brendan Tidd$^{1,4}$, Akansel Cosgun$^{2}$, J\"{u}rgen Leitner$^{1,3}$ and Nicolas Hudson$^{4}$
\thanks{$^{1}$Queensland University of Technology (QUT), Australia.}%
\thanks{$^{2}$Monash University, Australia}%
\thanks{$^{3}$LYRO Robotics Pty Ltd, Australia}%
\thanks{$^{4}$Robotics and Autonomous Systems Group, CSIRO, Pullenvale, QLD 4069, Australia}%
}

\begin{document}

\maketitle
\thispagestyle{empty}
\pagestyle{empty}

\begin{abstract}

The U.S. Defense Advanced Research Projects Agency (DARPA) Subterranean Challenge requires teams of robots to traverse difficult and diverse underground environments. Traversing small gaps is one of the challenging scenarios that robots encounter. Imperfect sensor information makes it difficult for classical navigation methods, where behaviours require significant manual fine tuning. In this paper we present a deep reinforcement learning method for autonomously navigating through small gaps, where contact between the robot and the gap may be required. We first learn a gap behaviour policy to get through small gaps (only centimeters wider than the robot). We then learn a goal-conditioned behaviour selection policy that determines when to activate the gap behaviour policy. We train our policies in simulation and demonstrate their effectiveness with a large tracked robot in simulation and on the real platform. In simulation experiments, our approach achieves 93\% success rate when the gap behaviour is activated manually by an operator, and 63\% with autonomous activation using the behaviour selection policy. In real robot experiments, our approach achieves a success rate of 73\% with manual activation, and 40\% with autonomous behaviour selection. While we show the feasibility of our approach in simulation, the difference in performance between simulated and real world scenarios highlight the difficulty of direct sim-to-real transfer for deep reinforcement learning policies. In both the simulated and real world environments alternative methods were unable to traverse the gap. 
\end{abstract}
\section{Introduction}
\label{sec:introduction}

\begin{figure}[t!]
    \captionsetup[subfigure]{labelformat=empty}
    \centering
    \subfloat[]{\includegraphics[trim=1cm 1cm 1cm 2cm, clip,width=0.45\columnwidth]{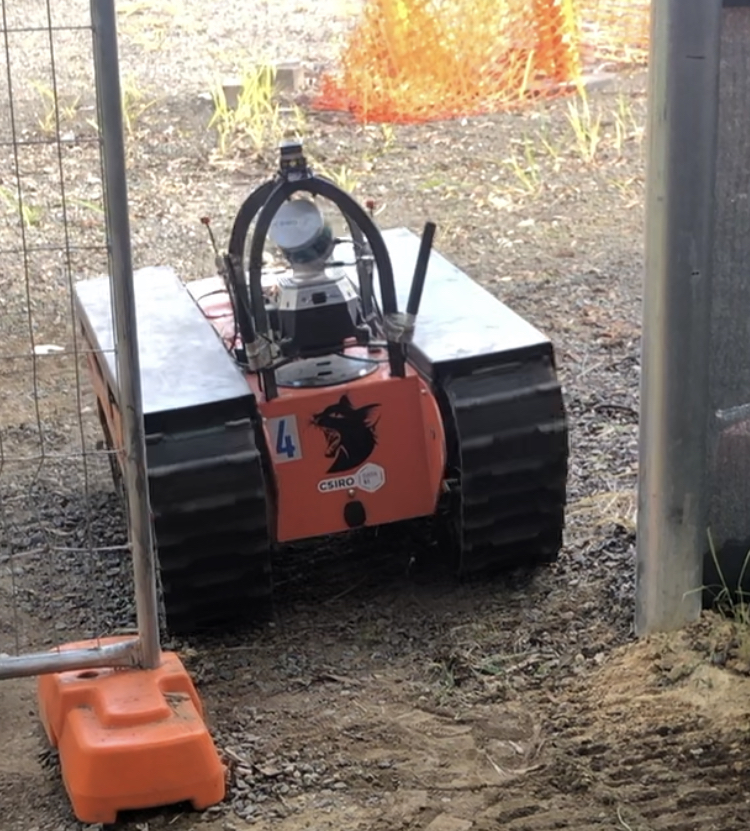}}
    \hfil
    \subfloat[]{\includegraphics[trim=5.0cm 0.0cm 0.2cm 15.3cm, clip, width=0.5\columnwidth]{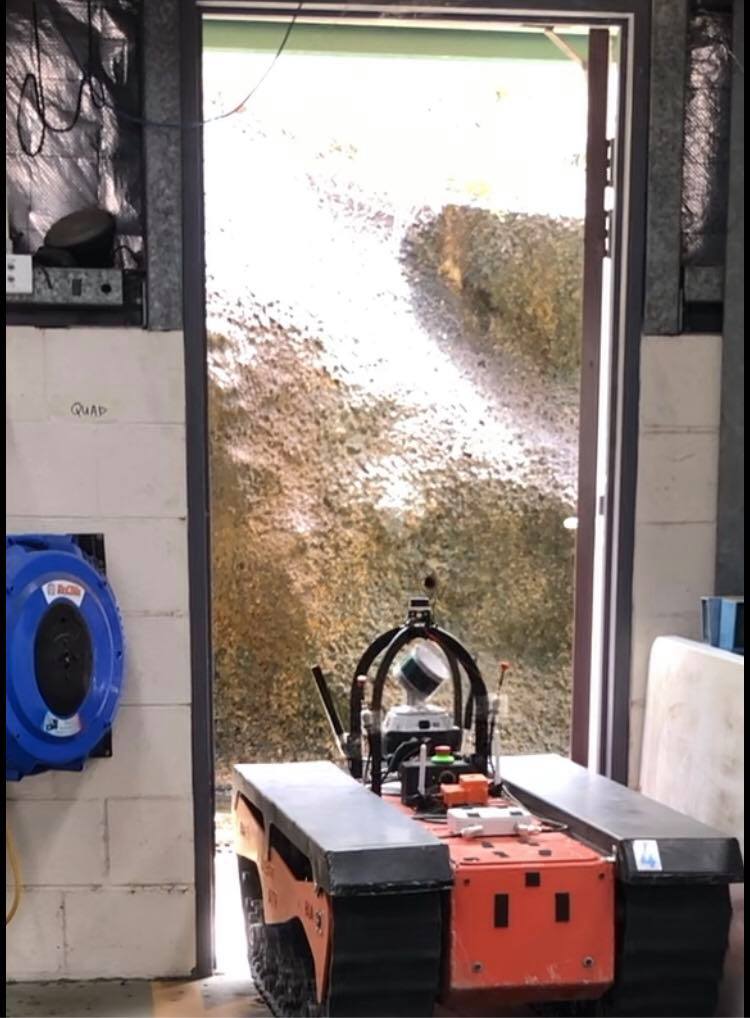}}
    \hfil
    \vspace*{-3mm}
    \caption{We study the problem of traversing narrow gaps with a mobile robot.}
    \label{fig:fig1}
\end{figure}

Navigation over various difficult terrain situations requires robust robotic platforms. The DARPA Subterranean Challenge (SubT)~\cite{darpa_darpa_2021} highlights the need for such platforms in unstructured underground circuits from tunnel, urban, and cave domains. Tracked robotic platforms are capable of navigating rough terrain and are well suited for subterranean search and rescue tasks, particularly where contact with the environment may be required. A number of the SubT scenarios requires the robots to navigate in human environments including small passages and doorways, where significant portions of the course may be obstructed by a narrow bottleneck. Fig.~\ref{fig:fig1} shows examples of narrow gaps centimeters wider than the OzBot BIA5 ATR robot~\cite{bia5_robotic_2021}, a large tracked robot. With teleoperation, a strategy often employed for doorways is to approach the gap on a slight angle, making contact with one side of the door, and then turn while in contact with the door. For autonomous operation, such a behaviour is difficult to derive manually, particularly for robots with imperfect velocity tracking, noisy perception systems and without sensors that can sense contact. Moreover, integrating such a behaviour into an existing behaviour stack requires understanding when to operate such a controller. 

In this paper, we propose a deep reinforcement learning approach to pass through narrow gaps, which is intended to be integrated with a classical behaviour stack. Reinforcement learning methods has lately been used to develop complex navigation behaviours and has improved the performance of navigation robots compared with classically designed controllers~\cite{rana_multiplicative_2020}. Our gap behaviour policy assumes the robot is already facing a gap, however, it is also important to know when to activate the gap behaviour, especially when other behaviours (e.g. path following) are used the majority of the time. Integration of a behaviour within a behaviour stack is necessary, though reinforcement learning methods typically act as a black box, with no way to know when a behaviour should be active. Learning when to use the gap behaviour so that it is autonomously activated is an important part of the problem.


Our contributions as follows:
\begin{itemize}
    \item A gap behaviour policy using deep reinforcement learning for controlling a mobile robot through small gaps
    \item A method to learn when to autonomously activate the gap behaviour policy
    \item Proof-of-concept demonstration of the effectiveness of the approach in simulation and in the real world, using a policy trained only in simulation
\end{itemize}


\section{Related Work}
\label{sec:related_work}

Designing an autonomous navigation system to handle difficult terrain scenarios is challenging. Often a collection of behaviours are required, where each behaviour has been designed for a unique situation~\cite{arkin_behavior-based_1998, hines_virtual_2021}. 

\subsection{Classical Methods for Robot Navigation}

Artificial potential fields (APF)~\cite{khatib_real-time_1986} is a widely used algorithm for navigation, where the robot is attracted to goal locations, and repelled from obstacles. Bug algorithms have also been used extensively~\cite{mcguire_comparative_2019}. The robot moves toward a goal unless an obstacle is encountered, then the robot moves along the boundary of the obstacle until it can once again move toward the goal. Small gaps is a limitation of these methods. Local minima with narrow gaps causes oscillations when the attractive and repelling forces are balanced~\cite{koren_potential_1991}. Probabilistic planners like Probabilistic Roadmap (PRM) and Rapidly-exploring Random Tree (RRT) also have difficulty with narrow spaces. Sun et al~\cite{zheng_sun_narrow_2005} introduce a bridge test to detect narrow passages for use with PRM planning.

Navigating through narrow gaps is challenging. Mujahed et al~\cite{mujahed_admissible_2018} proposed an admissibility gap: a virtual gap that satisfies the kinematic constraints of the vehicle to plan through tight spaces. In urban environments, Rusu et al ~\cite{rusu_laser-based_2009} design a geometric door detection method using 3D point cloud information, provided the door is within a set of specifications. A PR2 robot is able to detect a door and its handles~\cite{meeussen_autonomous_2010}, and negotiate the open or closed configuration to get through~\cite{chitta_planning_2010}. Cosgun~\cite{cosgun_context_2018} detects doors from door signs, and directs a robot through with pointing gestures.

Hines et al~\cite{hines_virtual_2021} designed a behaviour stack where each behaviour has a priority, and an admissibility criteria. A behaviour becomes active if it has a higher priority than the current behaviour, and is admissible. Admissibility criteria, and the behaviours themselves, are manually designed. The primary path follow method uses Hybrid A*~\cite{dolgov_path_2010} to plan a path through a costmap to a goal, considering the vehicle constraints. 

In each of these approaches, the robot does not make contact with the environment. Therefore gaps that appear smaller than the robot footprint are not considered for traversal, even if the robot may be able to get through. Sensory errors and coarse perception maps contribute to the failure of these methods. Furthermore, classical navigation methods often rely on manual tuning of parameters~\cite{khatib_real-time_1986, hines_virtual_2021, mcguire_comparative_2019}. Next, we review learning methods for navigation, aimed to reduce manual tuning.

\subsection{Learning Methods for Robot Navigation}

Learning methods used for navigation have shown promising results for robot navigation in human domains~\cite{rana_multiplicative_2020,bansal_combining_2019,kumar_learning_2019,moreno_automatic_2020,gupta_cognitive_2019} and in the field~\cite{kahn_badgr_2020}. 

Rana et al~\cite{rana_multiplicative_2020} efficiently learn a navigation policy from an APF prior controller while also learning a confidence for when the learned behaviour should be active, or when the robot should use the prior controller. 

With learning methods it is possible to navigate on various surfaces~\cite{kahn_badgr_2020}, and through narrow gaps~\cite{bansal_combining_2019} from RGB images. BADGR by Kahn et al~\cite{kahn_badgr_2020} learns a navigation policy directly from images with a self supervised approach. This method not only learns to navigate to a waypoint, but also to favour traversing a concrete path over grass. Bansal et al~\cite{bansal_combining_2019} learn to navigate through narrow doorways to get to a specified goal location using a monocular camera mounted to a robot.

Other learning approaches for robot navigation map and plan~\cite{gupta_cognitive_2019}, generate waypoints~\cite{kumar_learning_2019}, or use motion primitives~\cite{moreno_automatic_2020}. Gupta et al~\cite{gupta_cognitive_2019} learn a map and how to plan to a goal from an egocentric map with end-to-end reinforcement learning. Kumar et al~\cite{kumar_learning_2019} learn to plan at a high level of abstraction (turn left, turn right) from video data. Moreno et al~\cite{moreno_automatic_2020} generate auxiliary waypoint locations (critical navigation points) at problematic narrow regions.

With the exception of Rana et al~\cite{rana_multiplicative_2020}, learning methods are do not typically integrate with classical methods. In our previous work we introduce the idea of developing behaviours separately~\cite{tidd_guided_2020}, then learning when they should be used, either with a supervised approach~\cite{tidd_learning_2021}, or through reinforcement learning~\cite{tidd_learning_2021-1}. For our problem we require a behaviour that can learn to perform complex maneuvers (such as make contact with a door), and can be integrated into an existing behaviour stack.

\section{Robotic Platform}
\label{sec:robotic_platform}
The `Titan' all terrain robot (ATR) from BIA5~\cite{bia5_robotic_2021} is a 90Kg tracked vehicle, with a length of 1.4m and a width of 0.78m. It is the primary autonomous platform in a heterogeneous fleet of robots (tracked, legged, aerial) used by team CSIRO Data61's~\cite{hudson_heterogeneous_2021} entry in the SubT Challenge systems track~\cite{darpa_darpa_2021}. The Titan moves with nonholonomic motion following forward and heading velocity commands, and can turn on the spot (skid-steer).

\textbf{Perception Stack:}
Like all robots in team CSIRO Data61, the Titan has a `Catpack' perception system equipped with a tilted spinning Velodyne Lidar, Inertial Measurement Unit (IMU), 4 RGB cameras, and 3 thermal cameras. Each robot performs Simultaneous Localisation and Mapping (SLAM) onboard, providing a height map that is used by local navigation behaviours for fully autonomous navigation. Goals are provided by a higher level planner, encouraging thorough exploration in the search for artifacts (the full specifications of the Catpack and system solution can be found in~\cite{hudson_heterogeneous_2021} and ~\cite{hines_virtual_2021}). 

\textbf{Behaviour Stack:}
Among several manual defined controllers, the primary navigation behaviour is a \textit{path follow} behaviour that takes a goal point and calculates a safe path using hybrid A* search that associates a continuous state with each grid cell (in opposed to discrete cell center in standard A* search), allowing the path to consider vehicle kinematic constraints~\cite{dolgov_path_2010, hines_virtual_2021}. An \textit{orientation correction} behaviour recovers the robot to a level position in the case of extreme roll or pitch, and a \textit{decollide} behaviour will perform a maneuver into a safe location when there is fatal cost within the robot footprint. There is also a teleoperation behaviour (\textit{Joy}) for situations the operator can manually control a robot (when the robot is within communication range with the ground control station). Each behaviour is ranked in order of priority (shown in Fig.~\ref{fig:fig2}, behaviour stack), where the admissible behaviour with the highest priority is activated. Each behaviour is responsible for detecting its own admissibility.

\section{Method}
\label{sec:method}

\begin{figure}[tb!]
\centering
\includegraphics[width=\columnwidth]{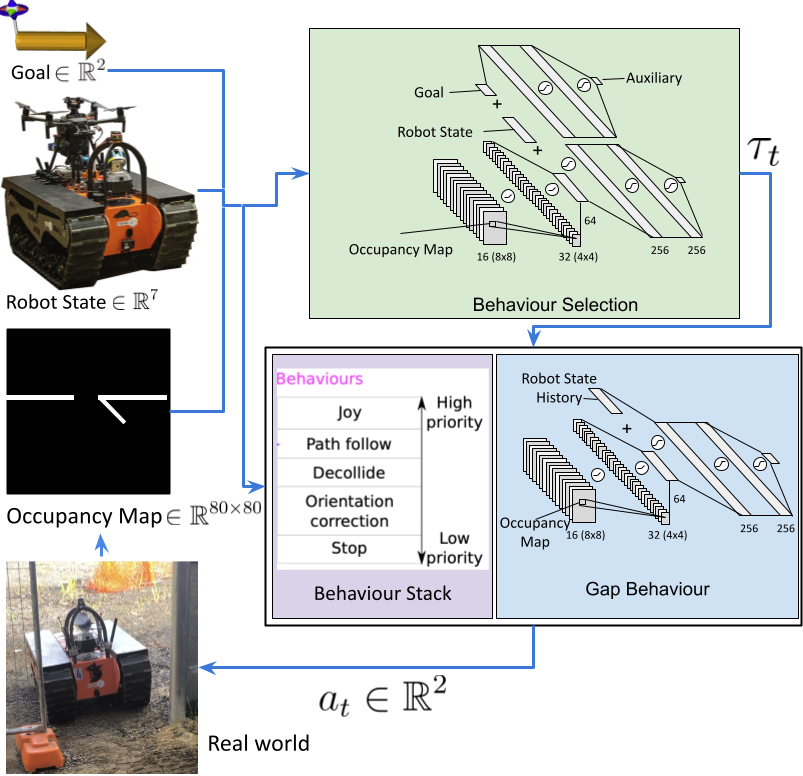}
\caption{We learn a gap behaviour and a behaviour selection policy using an occupancy map. The behaviour selection policy determines when the gap behaviour should be active based on a goal location, or when we should rely on a manually designed behaviour stack~\cite{hudson_heterogeneous_2021}. The system is trained in simulation, then tested on the real platform.}
\label{fig:fig2}
\end{figure}


Fig.~\ref{fig:fig2} outlines our system. The gap behaviour policy receives $s_t = [rs_t, I_t]$ (defined below), and outputs action $a_t$, consisting of forward and angular velocity about the $z$ axis (heading). The behaviour selection policy is also provided with $s_t$, as well as a goal location, and outputs switching parameter $\tau$. The gap behaviour should be activated by the behaviour selection policy when a goal is located behind a small gap, otherwise the behaviour stack should be active.

\subsection{Problem Description}
\label{sec:problem}

Traversing small gaps is important for underground search and rescue scenarios, such as those found in the SubT Challenge~\cite{darpa_darpa_2021}. The SubT Challenge involves detecting and localising artifacts (survivors, backpacks, helmets, etc) in underground environments (mine, urban, cave), where large portions of the course may only be accessible after a narrow constriction. Due to the large scale of the scenarios encountered, teams of robots are required to enter the underground environment, with a limitation that there can only be a single human operator. The human operator can command the robots through any method, though communications to robots are severely limited. Furthermore, coordinating multiple robots while reporting artifact detections in a timed competition (1 hour) presents a significant load on the operator. The robots must be as autonomous as possible. In these search and rescue scenarios the robot is allowed to make contact with the environment, but not intentionally damage it. In this paper we target traversing small doorways.

\subsection{Reinforcement Learning}
\label{sec:rl}

We consider our problem to be a Markov Decision Process $\text{MDP}$, where each state contains all relevant information to make a decision. An MDP is defined by tuple $\{\mathcal{S},\mathcal{A}, R, \mathcal{P}, \gamma\}$ where $s_t \in \mathcal{S}$, $a_t \in \mathcal{A}$, $r_t \in R$ are state, action and reward observed at time $t$, $\mathcal{P}$ is an unknown transition probability from $s_t$ to $s_{t+1}$ taking action $a_t$, and applying discount factor $\gamma$. 

With reinforcement learning, the goal is to maximise the sum of future rewards $R = \sum_{t=0}^{T}\gamma^tr_t$, where $r_t$ is the reward at time $t$. Actions are continuous values, sampled from a deep neural network ($a_t\sim\pi_\theta(s_t)$ for the gap behaviour policy, $\tau_t\sim\pi_\phi(s_t, goal)$ for the behaviour selection policy), and $s_t$ includes the robot state $rs_t$ and occupancy map $I_t$ (introduced below). We update each policy with Proximal Policy Optimisation (PPO)~\cite{schulman_proximal_2017}, using the implementation by OpenAI~\cite{dhariwal_openai_2017}.

 
\textbf{Robot State $rs_t$:} The robot state is a vector of 7 numbers containing the heading velocity $v_x$, and the angular position and velocities: $roll, pitch, yaw, v_{roll}, v_{pitch}, v_{yaw}$. Velocity terms are calculated with clipped finite differences. These terms are important for determining when the robot has made contact with the wall.

\textbf{Occupancy Map $I_t$:} A binary occupancy map with dimensions $[80,80,1]$ and resolution of 0.1m is provided to the each policy (gap behaviour policy, and behaviour selection policy). A value of 0 indicates a known and unoccupied cell, and a value of 1 indicates an unknown or occupied cell. The occupancy map moves with the $x$, $y$ translation and $yaw$ rotation of the robot, with the center of the robot located at the centre of the occupancy map. During training of both policies, the occupancy map is created using the ground truth position of the door frames and the robot location from the simulation. During testing (both in simulation and on the real robot), the occupancy map is supplied by the perception stack outlined above (Sec.~\ref{sec:robotic_platform}).

\begin{figure}[tb!]
\centering
\subfloat[Gap Behaviour]{\includegraphics[trim=1.8cm 3.6cm 1.8cm 3cm, clip, width=0.45\linewidth]{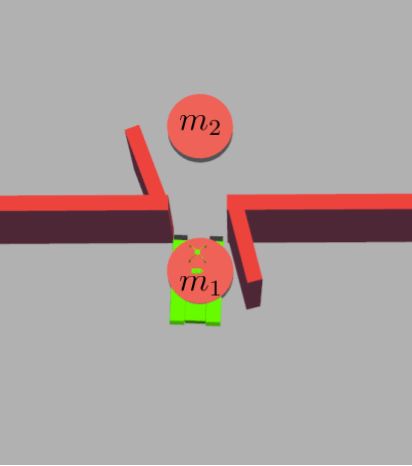}}
\hspace{0.3cm}
\subfloat[Behaviour Selection]{\includegraphics[trim=4.2cm 4.4cm 4cm 3.7cm, clip, width=0.41\linewidth]{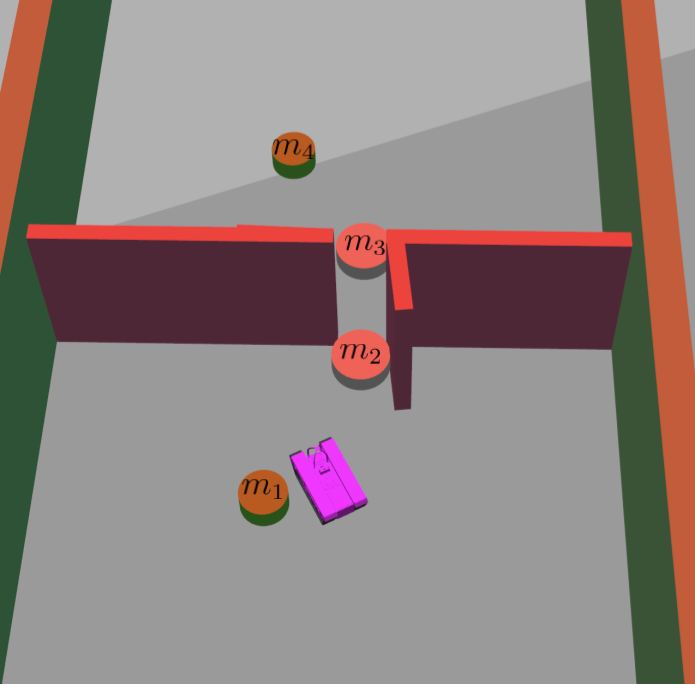}}

\caption{A simulation of a narrow door. The circular markers represent reward targets. a) The two doorway points are used by the gap behaviour policy. b) Behaviour selection policy uses two pairs of markers.}
\label{fig:doors}
\end{figure}

We train a gap behaviour policy to go through small gaps, and a behaviour selection policy that activates the gap behaviour when a goal appears behind a gap. The training setup is show in Fig.~\ref{fig:doors}. Fig.~\ref{fig:doors}a is used to train the gap behaviour policy, markers depict the entrance and exit to a gap ($m_1,m_2$). Fig.~\ref{fig:doors}b shows the setup for training the behaviour selection policy where additional markers show the goal points provided to behaviour selection ($m_1, m_4$).

\subsection{Gap Behaviour Policy}
\label{sec:behaviour}
The gap behaviour receives a stacked robot state message from the previous 5 timesteps: $[rs_t, rs_{t-1}, rs_{t-2}, rs_{t-3}, rs_{t-4}]$. The robot starts from a random initial position in front of a doorway, where the initial position is uniformly selected from a circle 1.5m back from the first marker $m_1$, with a radius of 1.5m. We also randomise the heading of the robot, choosing uniformly between -1.5 and 1.5 radians from the heading perpendicular to the gap. The target marker is set to the initial marker at the entrance to the gap $m_1$. All initial positions and heading ranges are selected based on the field of view of the perception system.

\textbf{Reward:} The reward terms encourage forward progress towards the target marker. Once the robot is within 0.5m of the target marker (initially set to marker $m_1$), the marker through the gap becomes the new target ($m_2$). This encourages the robot to approach the gap with the correct heading to make it through. The reward is defined as:
\[r_t = 1.5 r_{\dot{v}_x} - 0.1r_\theta - 0.05r_a - 0.75r_{\phi} - 0.001r_{s} - 0.25r_{v_{-x}}\]

where $r_{\dot{v}_x}$ is the velocity towards the target marker, $r_\theta$ is the error in heading between the robot and the target marker, $r_a$ is a penalty on angular velocity, $r_{\phi}$ is a penalty on roll and pitch angles, $r_{s}$ is a smoothing cost (previous actions should be close to the current actions), and $r_{v_{-x}}$ is a penalty for going backward. 

We train for 1000 episodes, where episodes are a fixed length of 180 steps (with a timestep of 0.1s), and the policy is updated at the end of an episode. 16 workers (CPU's) run in parallel, performing synchronised policy updates. Each robot is randomly inserted at the beginning of each episode, choosing between 98 different doorways (each with small variations in gap width, and door position). An episode is considered successful if the robot reaches $m_2$ (the marker after the door). All training parameters, including reward coefficients, are tuned empirically using task success.  


\subsection{Behaviour Selection Policy}
\label{sec:behaviour_selection_method}

While part of the problem we are considering involves controlling the robot through a narrow gap, to integrate the behaviour into the SubT behaviour stack, we must also know when the gap behaviour should be activated.

We once again use reinforcement learning to train an admissibility prediction that detects when the gap behaviour should be run, and when we should default to the behaviour stack (introduced in Sec:\ref{sec:robotic_platform}). The behaviour selection policy receives the robot state message $rs_t$, occupancy map $I_t$, and a goal location (distance and angle in the robot frame).

As we should only traverse a gap if there is a goal on the other side of the gap, we condition behaviour selection on having a goal. If the goal is on the same side of the gap as the robot, we do not want to activate the gap behaviour, but if is on the the other side we do. We have four markers $[m_1, m_2, m_3, m_4]$, shown in Fig.~\ref{fig:doors}b). Markers 1 and 4 are considered goals, and are provided to the behaviour selection policy. These goals would be provided by a higher level exploration planner. Markers 2 and 3 are the same as for the gap behaviour (entrance and exit of the gap) and are used to design the reward.

We start the robot from a random position sampled uniformly from a circle 4.5m before the door marker $m_2$ with a radius of 1.5m. $m_1$ and $m_4$ are also randomly selected from a uniform distribution, $m_1$ is located 1.5m \textit{before} $m_2$, and $m_4$ is located 1.5m \textit{after} $m_3$. The robot heading is randomly sampled from $(-\pi, \pi)$. This heading range is larger than the range used to train the gap behaviour, allowing the behaviour selection policy to be robust to any robot heading, where the gap behaviour is only expected to operate when the gap is in front of the robot.

\textbf{Reward:}
The reward term for behaviour selection is dependent on the current target marker. Once the robot is within 0.5m of a marker, we update the target marker to the next marker in the sequence. The robot should activate the behaviour stack (path follow behaviour) if it has not yet reached $m_1$, or has reached $m_3$. If the robot has reached $m_1$, but has not yet reached $m_3$, it should use the gap behaviour. The reward for behaviour selection is defined as:
\begin{equation}
\label{sup_reward}
    r_t = 
\begin{cases}
    1 - (\tau_t - 0.0)^2,& \text{if the target is } m_1 \text{ or } m_4 \\
    1 - (\tau_t - 1.0)^2,& \text{if the target is } m_2 \text{ or } m_3
\end{cases}
\end{equation}

Where $\tau$, a binary variable, is the output of the behaviour selection policy at time $t$. $\tau$ of 1.0 indicates activating the gap behaviour, and 0.0 the behaviour stack.


\textbf{Auxiliary task:} Learning an auxiliary task can help with the learning of visual features~\cite{jaderberg_reinforcement_2016}. We learn an auxiliary task with the behaviour selection policy. As shown in Fig.~\ref{fig:fig2}, the auxiliary task shares the convolution layers with behaviour selection, but has two separate fully connected layers. The auxiliary task uses a supervised learning loss to learn the gap width, and location of the gap markers. The auxiliary output is a vector of 5 terms: 
\[aux = [w, m_{2_{dist}}, m_{2_{angle}}, m_{3_{dist}}, m_{3_{angle}}] \]

Where $w$ is gap width, $dist$ and $angle$ are polar coordinates from the respective gap marker ($m_2, m_3$) to the robot ($dist$ is the distance to the robot, $angle$ is the heading error). We train the auxiliary term with mean squared error loss from the ground truth provided by the simulation. Learning the location of the gap has the additional benefit of providing visual information to the operator. If the gap behaviour is activated, the operator can clearly see the gap.


The gap behaviour policy and the behaviour selection policy are trained in simulation using the Gazebo simulator. We insert walls into the simulator so the robot must drive through gaps (sampled uniformly from $0.8 - 0.85$m). We also add doors in various open configurations. Due to the computational demands of running the full perception stack on a single computer while running multiple agents, the occupancy map is generated from simulation data while training. When the policy is being trained, the occupancy map is generated by inserting blocks into known locations in a global occupancy map, then sampling a robot-centric map using the robot pose ($x,y,yaw$). For testing, we rely on the occupancy map generated by the perception stack, for both simulation and real robot experiments (Sec.~\ref{sec:experiments}.B and Sec.~\ref{sec:experiments}.C).

\section{Experiments}
\label{sec:experiments}

\begin{figure}[t!]
    \vspace{-5mm}
    \centering
    \subfloat[Gap behaviour policy]{\includegraphics[trim=0cm 0.0cm 0cm 0cm, clip, width=0.5\columnwidth]{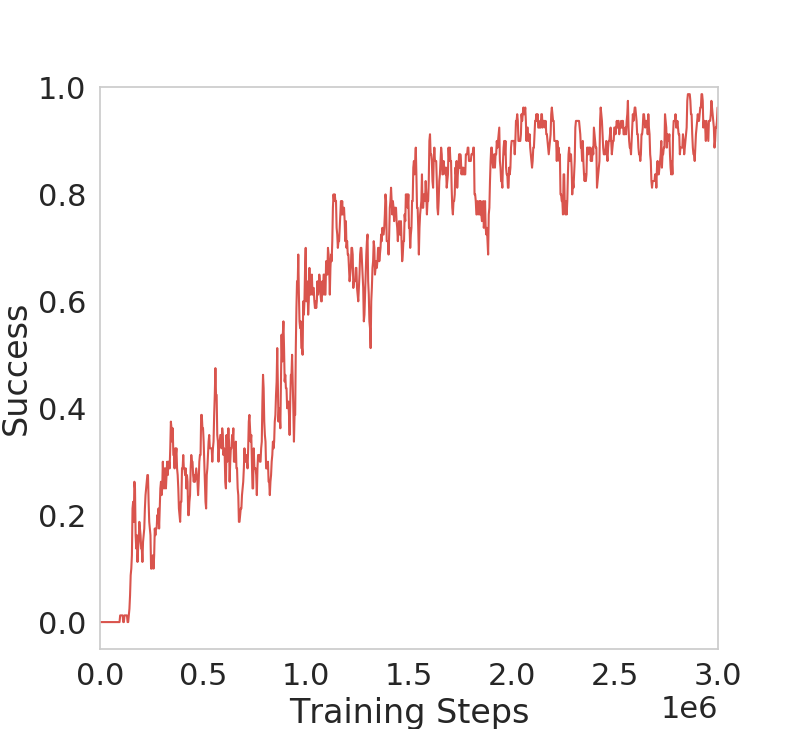}}
    \hfil
    \subfloat[Behaviour selection policy]{\includegraphics[trim=0cm 0cm 0cm 0cm, clip,width=0.5\columnwidth]{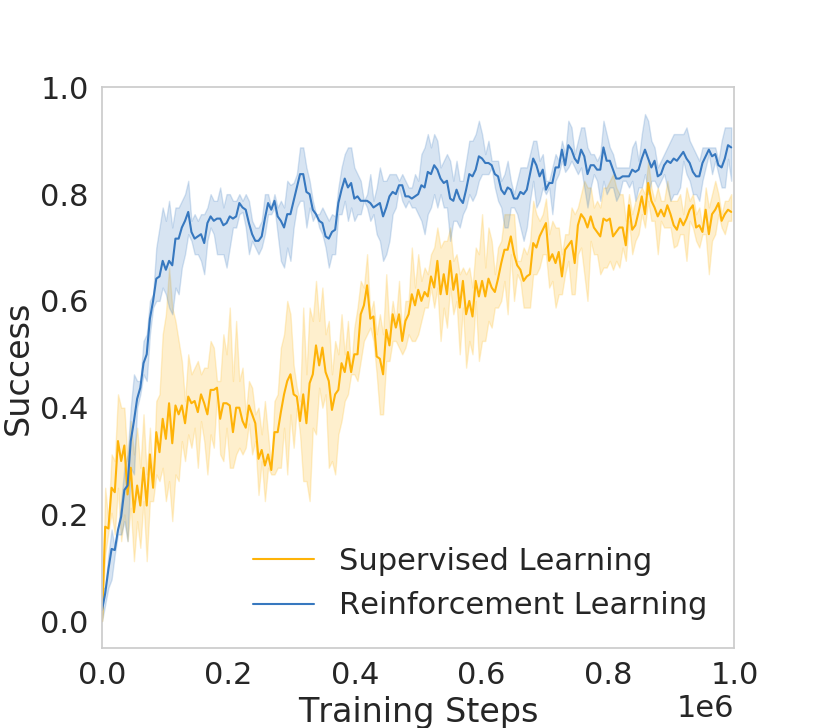}}
    \hfil
    \caption{Success rates during training for the gap behaviour policy and the behaviour selection policy.}
    \label{fig:learning_curves}
\end{figure}


We first show the results of training our policies (Sec.~\ref{sec:training}). Sec.~\ref{sec:simulation} shows the results in simulation using the perception stack (introduced in ~\ref{sec:problem}), and the real robot results are shown in Sec.~\ref{sec:real}. Fig~\ref{fig:modes}.a and b) show the setup for the simulated and real doors.

\subsection{Training Results}
\label{sec:training}

The \textbf{gaps behaviour policy} achieves a success rate of 93.8\% on the simulated environment (98 gaps of random width and door type). The success rate while training is shown in Fig.~\ref{fig:learning_curves}.a). The behaviour performs several complex maneuvers that would be difficult to encode manually. These maneuvers include approaching the gap with one corner and making contact with the door, reversing and re-approaching when the robot pitches, and `wiggling' when the robot is part way through but not making progress.

Once the gap behaviour policy is trained, we train the \textbf{behaviour selection policy} with the method from Sec\ref{sec:behaviour_selection_method}. The results shown in Fig~\ref{fig:learning_curves}.b), display the success rate for behaviour selection training. We compare using a supervised loss to using our reinforcement learning method for learning behaviour selection:

~\textbf{Supervised Learning:} Behaviour selection could be considered a supervised learning problem where robot state with occupancy map at each timestep are taken as input, and are given a label. We use labels of 0 if the target marker is $m_1$ or $m_4$, and 1 if the target marker is $m_2$ or $m_3$. We update network weights using mean squared error loss.

~\textbf{Reinforcement Learning:}
We train the behaviour selection policy using the reinforcement learning approach introduced in Sec~\ref{sec:behaviour_selection_method}, where we use a reinforcement learning update with the reward from Equation~\ref{sup_reward}.

In both examples we also predict the auxiliary terms, which include door size, and marker locations relative to the robot. Both methods were trained with online data using the same training parameters. As can be seen in Fig.~\ref{fig:learning_curves}, the \textbf{reinforcement learning} approach achieves more successful door traversals, with an asymptotic success rate of 88.7\%, compared to 75.8\% with the \textbf{supervised learning} approach. We attribute the difference between the methods on the reinforcement learning approach considering the sum of future behaviour selections, whereas the supervised approach only considers the current timestep. We found that excluding the auxiliary terms from training had very little effect on success rate. We also tested various additional parameters in the reward function but found almost no effect on success rate. We have omitted these results from Fig.~\ref{fig:learning_curves}b) for clarity. 




\subsection{Simulation Results}
\label{sec:simulation}

We test our gap behaviour and behaviour selection policies in simulation on an unseen test world, using the ~\textit{perception stack} (Lidar scans are used to produce the occupancy map). The test world has gaps uniformly sampled from $0.8-0.85$m, with random door types. We compare several modes of operation including using the behaviour stack without the gap behaviour, manually activating the gap behaviour, and using the behaviour selection policy to autonomously activate the gap behaviour.

~\textbf{Behaviour Stack:} The current navigation stack used by the Titan platform uses several behaviours described in Sec~\ref{sec:problem}: path follow, orientation correction and decollide. Path follow is the dominant behaviour where hybrid A* search is used to find a path to a local goal. For more details about the behaviour stack see~\cite{hines_virtual_2021} and~\cite{hudson_heterogeneous_2021}. For these experiments the operator inserts a waypoint on the opposite side of a gap.

~\textbf{Manual Gap Behaviour Selection:} The operator can manually call the gap behaviour from a GUI. The operator must first insert a waypoint so the robot approaches the gap (running the behaviour stack), then activate the gap behaviour, and when the robot is through the gap once again activate the behaviour stack to get the robot to the waypoint after the door.

~\textbf{Autonomous Gap Behaviour Selection:} The gap behaviour is autonomously selected when the conditions are met (there is a goal behind a small gap). The operator can place a waypoint on the other side of a door and behaviour selection will activate the behaviour stack or the gap behaviour autonomously.

\begin{table}[h!]
    \centering
    \begin{adjustbox}{max width=1.0\columnwidth}
    \begin{tabular}{cccccc}
                                      & Success \% & Time(s) & Operator Actions \\
       \hline
        Behaviour Stack      & 0.0            &  50.2 &    1  \\
        Manual Gap Behaviour Selection        & 93.3          &  29.7 &    3  \\
        Auto Gap Behaviour Selection  & 63.3          &  45.3 &    1 \\
    \hline
    \end{tabular}
    \end{adjustbox}
    \caption{Simulation results for passing through a narrow gap (30 trials)}
    \label{tab:sim_results}
\end{table}


We perform 30 trials and present results in Table~\ref{tab:sim_results}. We show the average success rate, time until the final waypoint, and the number of operator actions. We can see that the behaviour stack cannot traverse the gap and reaches a timeout (no progress after 30 seconds). Manually activating the gap behaviour has a high success rate (93.3\%), and autonomously selecting behaviours has a reasonable success rate of 63.3\%. Of the failure cases with behaviour selection, the most common cause was the behaviour selector not recognising the gap. We attribute this failure to distribution mismatch between the perception used during training, and the noise introduced by the ~\textit{perception stack}. Other failure cases include the gap behaviour failing, and several cases where oscillation between the gap behaviour and the decollide behaviour caused a timeout on the waypoint. Where decollide did not activate (behaviour selection did not switch back to the behaviour stack midway through the gap), the average time through the gap was 17.3 seconds, suggesting behaviour selection is a viable method, however requires further investigation to ensure robustness.

\begin{figure}[t!]
    \centering
    \subfloat[Gazebo Simulator]{\includegraphics[trim=3cm 7cm 3cm 6cm, clip, width=0.5\linewidth]{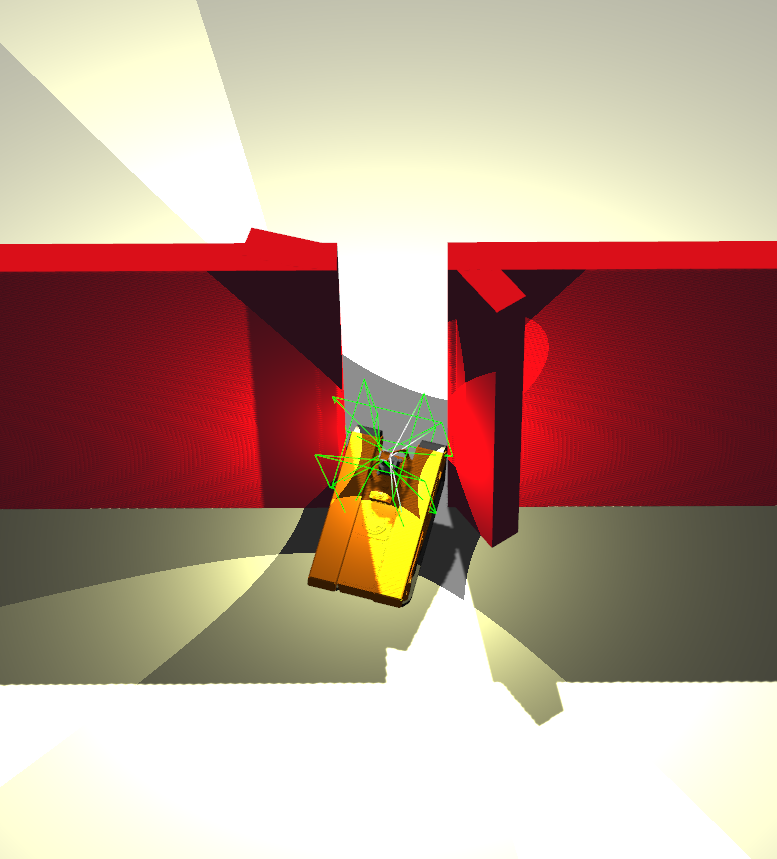}}
    \hspace{0.3cm}
    \subfloat[Real Robot]{\includegraphics[trim=3cm 4cm 0cm 12.2cm, clip, width=0.4\linewidth]{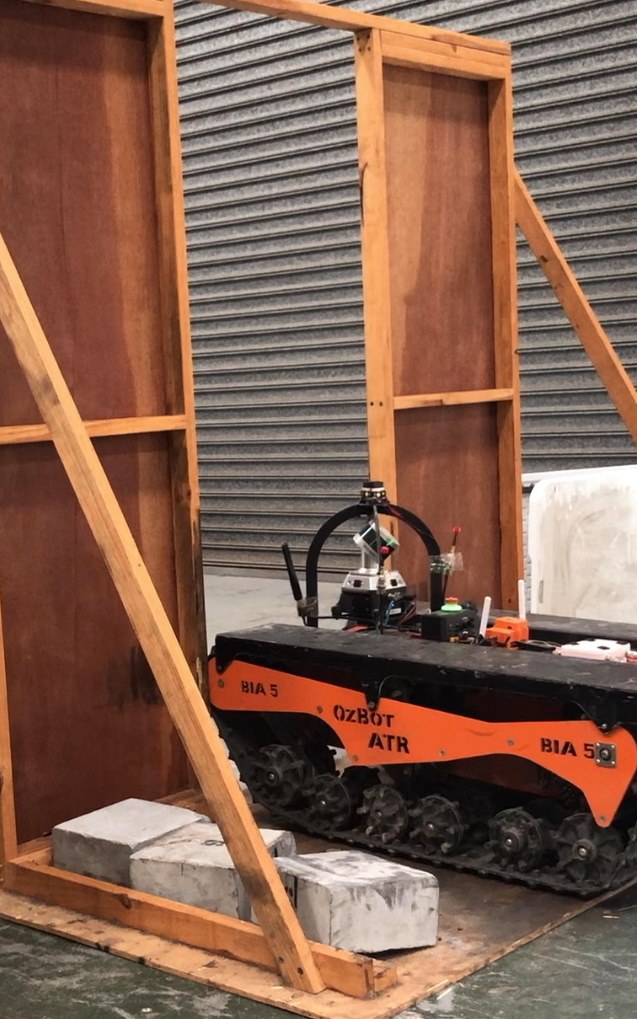}}
    \caption{Experimental setup in simulation and with the real robot.}
    \label{fig:modes}
\end{figure}

\subsection{Real Robot Results}
\label{sec:real}


\begin{table}[h!]
    \vspace{-2mm}
    \centering
    \begin{adjustbox}{max width=1.0\columnwidth}
    \begin{tabular}{cccccc}
                                      & Success \% & Time(s) & Operator Actions \\
       \hline
        Behaviour Stack              & 0.0           &  44.6 &    1  \\
        Manual Gap Behaviour Selection               & 73.3          &  24.1 &    3  \\
        Auto Gap Behaviour Selection         & 40.0          &  30.9 &    1 \\
    \hline
    \end{tabular}
    \end{adjustbox}
    \caption{Real robot results for passing through a narrow gap (15 trials).}
    \label{tab:real_results}
\end{table}

Table~\ref{tab:real_results} show our final experiments applied in a door scenario on a real robot. We use a constructed wooden door rather than a real door as the gap behaviour is allowed to contact the door, which can result in the robot damaging the environment. The constructed door shown in Fig~\ref{fig:modes}.b).

We run 15 trials for all experiments, results are shown in Table~\ref{tab:real_results}. We confirm our behaviour succeeds in more than 70\% of trials, where the current behaviour stack cannot traverse the gap, timing out after 30 seconds. The behaviour selection method is successful 40\% of the time. Both manually evoking the gap behaviour, and relying on autonomous behaviour selection had the same failure case: the edge of the robot would grab the protrusion of the door and the tracks would spin on the slippery surface. This was not seen in the simulation trials as track slippage is not modeled well in simulation. While this situation may not occur regularly with real doors, future policies will be trained with examples of gaps with small protruding edges. 

Autonomous behaviour selection had a much lower success rate than manually activating the gap behaviour. This is due to the late switching from the behaviour stack to the gap behaviour, and the limitation of our test door (protruding edges). With late switching, the gap behaviour was more likely to get the robot caught on the outer wooden structure of the door. Future behaviour selection models would encourage switching to the gap behaviour earlier, allowing more time to align with the door.

In one case with behaviour selection, the robot was mostly through the gap when the behaviour switched back to the behaviour stack, triggering a decollide behaviour and causing the robot to go in and out of the gap. Training the behaviour selection policy with the marker after the door ($m_3$) further back should result in the gap behaviour being active until after the robot has cleared the door.

While these limitations must be addressed before the behaviour can be properly integrated into the behaviour stack for autonomous operation, the manual activation of the gap behaviour policy has been utilised in a SubT 1 hour test run with 5 different robots (3 Titan's, 1 Ghost Vision60 quadruped, and a drone). A small gap was identified when the behaviour stack timed out at a constriction in the tunnel. The operator manually intervened to execute the gap behaviour policy, resulting in the robot successfully going through the gap.



\section{Conclusion}
\label{sec:conclusion}

We present a reinforcement learning method for developing a gap behaviour policy to traverse small gaps. We also learn a behaviour selection policy that determines when to autonomously activate the gaps behaviour (when a goal is through a small gap). We show that our gap behaviour policy is more reliable than the current stack of tuned behaviours, both in simulation (93\% compared to 0\%) and on a real robot (73\% compared to 0\%), though we recognise that the autonomous behaviour selection needs to be more robust before it can be relied upon for fully autonomous integration (success rates of 63\% in simulation and 40\% in real trials). We have identified a limitation of transferring from the simulation to the real robot by observing the way the robot gets stuck on a protruding door frame. Future work will look at improving the reliability of the system for a wider range of gap scenarios, including updating the simulation to include these cases, and adding noise to the occupancy map during training to mimic the noisy output of the real perception stack. 

A limitation of our method occurs when gaps are smaller than the robot. It is likely in these situations the gap behaviour policy will still be selected for operation and try to traverse the gap. Methods to handle these scenarios will be considered in future work.


\section*{ACKNOWLEDGMENT}
This research was, in part funded by the US Government under the
DARPA Subterranean Challenge. The views, opinions, and findings expressed are those of the authors and should not be interpreted as representing the official views or policies of the Department of Defense or the U.S. Government. Approved for Public Release, Distribution Unlimited.



\addtolength{\textheight}{-2cm}   
\bibliographystyle{IEEEtran}
\bibliography{references}

\begin{thebibliography}{10}
\providecommand{\url}[1]{#1}
\csname url@samestyle\endcsname
\providecommand{\newblock}{\relax}
\providecommand{\bibinfo}[2]{#2}
\providecommand{\BIBentrySTDinterwordspacing}{\spaceskip=0pt\relax}
\providecommand{\BIBentryALTinterwordstretchfactor}{4}
\providecommand{\BIBentryALTinterwordspacing}{\spaceskip=\fontdimen2\font plus
\BIBentryALTinterwordstretchfactor\fontdimen3\font minus
  \fontdimen4\font\relax}
\providecommand{\BIBforeignlanguage}[2]{{%
\expandafter\ifx\csname l@#1\endcsname\relax
\typeout{** WARNING: IEEEtran.bst: No hyphenation pattern has been}%
\typeout{** loaded for the language `#1'. Using the pattern for}%
\typeout{** the default language instead.}%
\else
\language=\csname l@#1\endcsname
\fi
#2}}
\providecommand{\BIBdecl}{\relax}
\BIBdecl

\bibitem{darpa_darpa_2021}
\BIBentryALTinterwordspacing
DARPA, ``{DARPA} {Subterranean} {Challenge},'' 2021. [Online]. Available:
  \url{https://www.subtchallenge.com/}
\BIBentrySTDinterwordspacing

\bibitem{bia5_robotic_2021}
\BIBentryALTinterwordspacing
BIA5, ``Robotic {Solutions},'' 2021. [Online]. Available:
  \url{https://bia5.com/robotics-solutions/.}
\BIBentrySTDinterwordspacing

\bibitem{rana_multiplicative_2020}
K.~Rana, V.~Dasagi, B.~Talbot, M.~Milford, and N.~Sünderhauf, ``Multiplicative
  {Controller} {Fusion}: {A} {Hybrid} {Navigation} {Strategy} {For}
  {Deployment} in {Unknown} {Environments},'' \emph{IEEE International
  Conference on Intelligent Robots and Systems}, 2020.

\bibitem{arkin_behavior-based_1998}
R.~C. Arkin, \emph{An {Behavior}-based {Robotics}}.\hskip 1em plus 0.5em minus
  0.4em\relax MIT Press, 1998.

\bibitem{hines_virtual_2021}
T.~Hines, K.~Stepanas, F.~Talbot, I.~Sa, J.~Lewis, E.~Hernandez, N.~Kottege,
  and N.~Hudson, ``\BIBforeignlanguage{en}{Virtual {Surfaces} and {Attitude}
  {Aware} {Planning} and {Behaviours} for {Negative} {Obstacle}
  {Navigation}},'' \emph{\BIBforeignlanguage{en}{arXiv preprint
  arXiv:2010.16018}}, Jan. 2021.

\bibitem{khatib_real-time_1986}
O.~Khatib, ``Real-{Time} {Obstacle} {Avoidance} for {Manipulators} and {Mobile}
  {Robots},'' \emph{IJRR}, 1986.

\bibitem{mcguire_comparative_2019}
K.~N. McGuire, G.~C. H.~E. de~Croon, and K.~Tuyls, ``A comparative study of bug
  algorithms for robot navigation,'' \emph{Robotics and Autonomous Systems},
  2019.

\bibitem{koren_potential_1991}
Y.~Koren and J.~Borenstein, ``Potential field methods and their inherent
  limitations for mobile robot navigation,'' 1991.

\bibitem{zheng_sun_narrow_2005}
{Zheng Sun}, D.~Hsu, {Tingting Jiang}, H.~Kurniawati, and J.~H. Reif, ``Narrow
  passage sampling for probabilistic roadmap planning,'' \emph{IEEE
  Transactions on Robotics}, 2005.

\bibitem{mujahed_admissible_2018}
M.~Mujahed, D.~Fischer, and B.~Mertsching, ``Admissible gap navigation: {A} new
  collision avoidance approach,'' \emph{Robotics and Autonomous Systems}, 2018.

\bibitem{rusu_laser-based_2009}
R.~Rusu, W.~Meeussen, S.~Chitta, and M.~Beetz, ``Laser-based perception for
  door and handle identification,'' \emph{Advanced Robotics}, 2009.

\bibitem{meeussen_autonomous_2010}
W.~e.~a. Meeussen, ``Autonomous door opening and plugging in with a personal
  robot,'' in \emph{{IEEE} {International} {Conference} on {Robotics} and
  {Automation}}, 2010.

\bibitem{chitta_planning_2010}
S.~Chitta, B.~Cohen, and M.~Likhachev, ``Planning for autonomous door opening
  with a mobile manipulator,'' in \emph{2010 {IEEE} {International}
  {Conference} on {Robotics} and {Automation}}, 2010.

\bibitem{cosgun_context_2018}
A.~Cosgun and H.~I. Christensen, ``Context-aware robot navigation using
  interactively built semantic maps,'' \emph{Paladyn, Journal of Behavioral
  Robotics}, vol.~9, no.~1, pp. 254--276, 2018.

\bibitem{dolgov_path_2010}
D.~Dolgov, S.~Thrun, M.~Montemerlo, and J.~Diebel, ``Path {Planning} for
  {Autonomous} {Vehicles} in {Unknown} {Semi}-structured {Environments},''
  \emph{The International Journal of Robotics Research}, 2010.

\bibitem{bansal_combining_2019}
S.~Bansal, V.~Tolani, S.~Gupta, J.~Malik, and C.~Tomlin, ``Combining {Optimal}
  {Control} and {Learning} for {Visual} {Navigation} in {Novel}
  {Environments},'' \emph{arXiv:1903.02531}, 2019.

\bibitem{kumar_learning_2019}
A.~Kumar, S.~Gupta, and J.~Malik, ``Learning {Navigation} {Subroutines} from
  {Egocentric} {Videos},'' \emph{arXiv:1905.12612}, 2019.

\bibitem{moreno_automatic_2020}
F.-A. Moreno, J.~Monroy, J.-R. Ruiz-Sarmiento, C.~Galindo, and
  J.~Gonzalez-Jimenez, ``Automatic {Waypoint} {Generation} to {Improve} {Robot}
  {Navigation} {Through} {Narrow} {Spaces},'' \emph{Sensors}, 2020.

\bibitem{gupta_cognitive_2019}
S.~Gupta, V.~Tolani, J.~Davidson, S.~Levine, R.~Sukthankar, and J.~Malik,
  ``Cognitive {Mapping} and {Planning} for {Visual} {Navigation},''
  \emph{arXiv:1702.03920}, 2019.

\bibitem{kahn_badgr_2020}
G.~Kahn, P.~Abbeel, and S.~Levine, ``{BADGR}: {An} {Autonomous}
  {Self}-{Supervised} {Learning}-{Based} {Navigation} {System},''
  \emph{arXiv:2002.05700}, 2020.

\bibitem{tidd_guided_2020}
B.~Tidd, N.~Hudson, and A.~Cosgun, ``Guided {Curriculum} {Learning} for
  {Walking} {Over} {Complex} {Terrain},'' in \emph{Australasian {Conference} on
  {Robotics} and {Automation}}, 2020.

\bibitem{tidd_learning_2021}
B.~Tidd, N.~Hudson, A.~Cosgun, and J.~Leitner, ``Learning {When} to {Switch}:
  {Composing} {Controllers} to {Traverse} a {Sequence} of {Terrain}
  {Artifacts},'' in \emph{{IEEE} {International} {Conference} on {Intelligent}
  {Robots} and {Systems}}, 2021.

\bibitem{tidd_learning_2021-1}
B.~{Tidd}, N.~Hudson, A.~Cosgun, and J.~Leitner, ``Learning {Setup} {Policies}:
  {Reliable} {Transition} {Between} {Locomotion} {Behaviours},''
  \emph{arXiv:2101.09391 [cs]}, Jan. 2021.

\bibitem{hudson_heterogeneous_2021}
N.~Hudson, F.~Talbot, M.~Cox, J.~Williams, T.~Hines, A.~Pitt, B.~Wood,
  D.~Frousheger, K.~L. Surdo, T.~Molnar, R.~Steindl, M.~Wildie, I.~Sa,
  N.~Kottege, K.~Stepanas, E.~Hernandez, G.~Catt, W.~Docherty, B.~Tidd, B.~Tam,
  S.~Murrell, M.~Bessell, L.~Hanson, L.~Tychsen-Smith, H.~Suzuki, L.~Overs,
  F.~Kendoul, G.~Wagner, D.~Palmer, P.~Milani, M.~O’Brien, S.~Jiang, S.~Chen,
  and R.~C. Arkin, ``\BIBforeignlanguage{en}{Heterogeneous {Ground} and {Air}
  {Platforms}, {Homogeneous} {Sensing}: {Team} {CSIRO} {Data61}’s {Approach}
  to the {DARPA} {Subterranean} {Challenge}},''
  \emph{\BIBforeignlanguage{en}{submitted to the DRC Finals Special Issue of
  the Journal of Field Robotics}}, 2021.

\bibitem{schulman_proximal_2017}
J.~Schulman, F.~Wolski, P.~Dhariwal, A.~Radford, and O.~Klimov, ``Proximal
  {Policy} {Optimization} {Algorithms},'' \emph{arXiv:1707.06347}, 2017.

\bibitem{dhariwal_openai_2017}
\BIBentryALTinterwordspacing
P.~Dhariwal, C.~Hesse, O.~Klimov, A.~Nichol, M.~Plappert, A.~Radford,
  J.~Schulman, S.~Sidor, Y.~Wu, and P.~Zhokhov, \emph{{OpenAI} {Baselines}},
  2017. [Online]. Available: \url{https://github.com/openai/baselines}
\BIBentrySTDinterwordspacing

\bibitem{jaderberg_reinforcement_2016}
M.~Jaderberg, V.~Mnih, W.~M. Czarnecki, T.~Schaul, J.~Z. Leibo, D.~Silver, and
  K.~Kavukcuoglu, ``Reinforcement {Learning} with {Unsupervised} {Auxiliary}
  {Tasks},'' in \emph{International {Conference} on {Learning}
  {Representations}}, 2016.

\end{thebibliography}
\end{document}